\documentclass[wcp]{jmlr}

\usepackage{longtable}
\usepackage{microtype}
\usepackage{csquotes}
\usepackage{booktabs}
\usepackage{multirow}
\usepackage{makecell}
\usepackage{array}
\usepackage{listings}
\usepackage{xcolor}
\usepackage{lineno}
\newcolumntype{Y}{>{\centering\arraybackslash}X} 

\SetKwInOut{Input}{Input}
\SetKwInOut{Output}{Output}
\SetKwComment{Comment}{$\triangleright$\ }{}

\usepackage{listings}
\usepackage{xcolor}

\usepackage{algorithm}
\usepackage{algorithmic}

\definecolor{promptred}{HTML}{8B1A1A} 

\lstdefinestyle{paperprompt}{
    basicstyle=\ttfamily\small\color{promptred},
    frame=single,                                
    rulecolor=\color{black},                    
    breaklines=true,                             
    breakatwhitespace=false,                     
    postbreak=\mbox{\textcolor{promptred}{$\hookrightarrow$}\space}, 
    columns=fullflexible,                        
    keepspaces=true,                              
    aboveskip=1em,                                
    belowskip=1em,
    captionpos=b
}

\makeatletter
\let\Ginclude@graphics\@org@Ginclude@graphics 
\makeatother

\jmlrproceedings{}{}

\title[SHAP-Guided Implicit-Trajectory Generation for Metadata-Free LLM-Based AutoFE]{SIGMA: SHAP-Guided Implicit-Trajectory Generation for Metadata-Free LLM-Based AutoFE}

 \author{
  \Name{Xuan Zheng} \Email{zheng-xuan-dz@ynu.jp}\\
  \Name{Kento Uchida} \Email{uchida-kento-fz@ynu.ac.jp}\\
  \Name{Shinichi Shirakawa} \Email{shirakawa-shinichi-bg@ynu.ac.jp}\\
\addr Yokohama National University}

\editors{}

\begin{document}

\makeatletter
\let \@jmlrpages \@empty
\makeatother
\maketitle

\begin{abstract}
  Recent research has leveraged Large Language Models (LLMs) to enhance Automated Feature Engineering (AutoFE) through semantic descriptions and trajectory-based prompting.
  However, there exist two challenges that limit their applicability and scalability in long-horizon optimization:
  (1) semantic metadata is unavailable in many practical settings, and 
  (2) trajectory accumulation increases the risk of exceeding the context window, while without it, the generation process can become unstable, leading to becoming stuck in the local optima and a high duplicate rate of generated features.
  To this end, we propose a \textbf{S}HAP-enhanced \textbf{I}mplicit-trajectory \textbf{G}eneration for \textbf{M}etadata-free \textbf{A}utoFE (SIGMA), a scalable constant-context optimization framework.
  SIGMA leverages SHAP values to provide task-aware signals for guiding group feature generation instead of semantic information.
  In addition, we adopt an \textbf{EX}posed-feature \textbf{I}mplicit \textbf{T}rajectory (EXIT) approach, where the exposed features in the prompt implicitly represent the trajectory.
  Empirical results demonstrate that SIGMA achieves performance comparable to the state-of-the-art (SOTA) LLM baselines with a nearly constant prompt length.
  Notably, EXIT significantly reduces the duplicate ratio of generated features from 37.2\% to 6.8\%.
  At the same time, SIGMA matches traditional SOTA performance with only 5.4 features on average, demonstrating substantial efficiency gains in feature utilization.
\end{abstract}

\begin{figure}[htb!]
  \centering
  \includegraphics[width=1.0\textwidth]{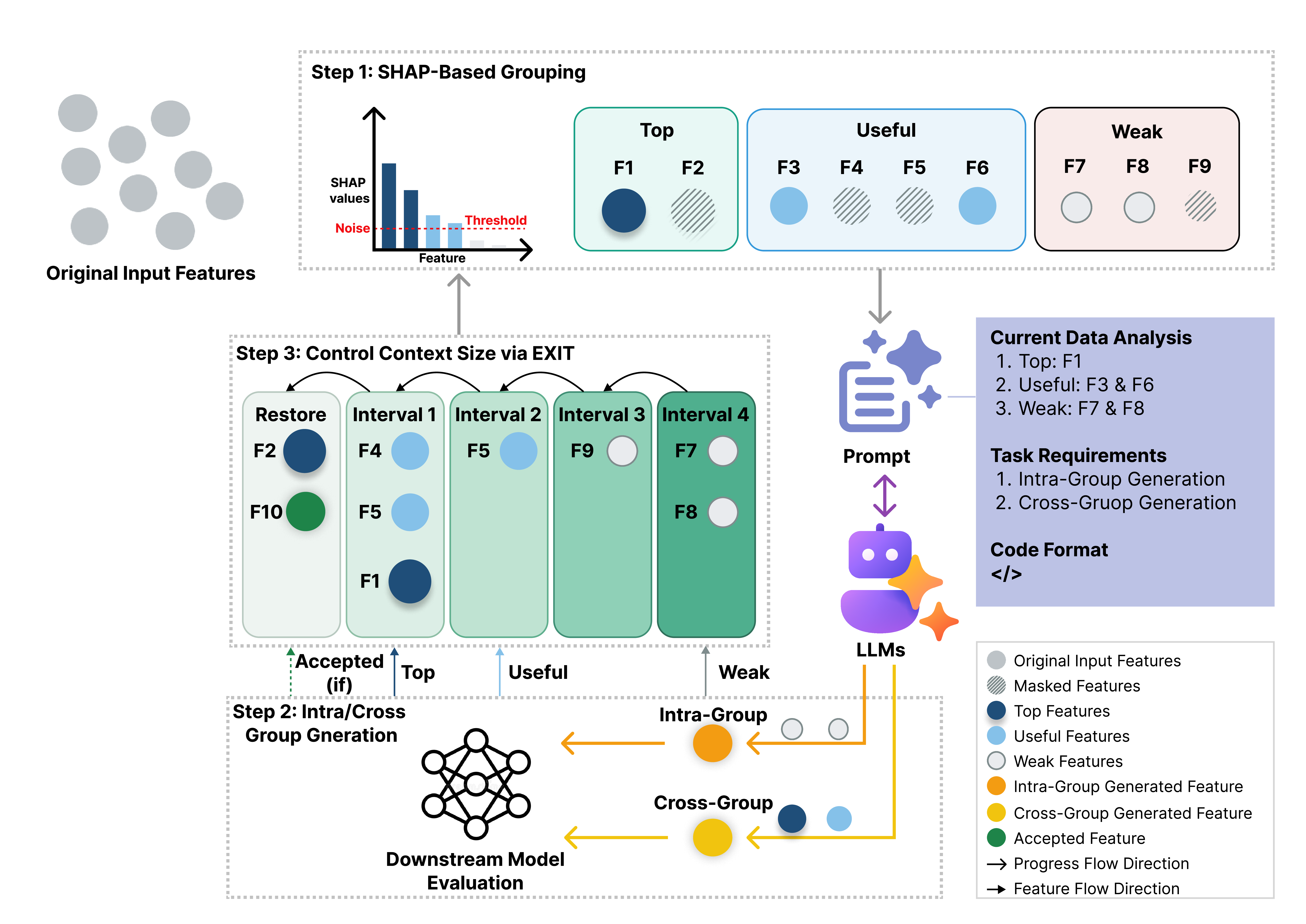}
  \caption{The overview of SIGMA.}
  \label{fig:overview}
\end{figure}

\section{Introduction}
Automated Feature Engineering (AutoFE) \citep{hutter2019automated} is an important component of AutoML \citep{ravishankar2025survey}.
A primary objective of AutoFE is generating new features to enhance the representational power of the original feature set.
It improves AutoML efficiency and robustness, and has been widely applied in domains such as finance and medicine~\citep{hollmann2023large,lucas2020towards,waring2020automated}.

Traditionally, AutoFE methods primarily adopt an expansion-reduction framework that explores large combinatorial spaces of predefined feature transformations \citep{zhang2023openfe,hollmann2023large}.
These approaches have demonstrated strong performance under sufficient search budgets.
However, the manually defined search space is not only complex to design but also limits the exploration, potentially leading to sub-optimal results \citep{abhyankar2025llm}.
In addition, it is hard to interpret the enormous number of generated features.

Since Large Language Models (LLMs) \citep{chang2024survey} have shown strong reasoning \citep{wei2022chain} and In-Context Learning (ICL) \citep{dong2024survey} capabilities, extensive research has explored leveraging LLMs to enhance AutoFE through sequential optimization.
By providing semantic (features and task) descriptions and statistical information \citep{fathollahzadeh2025demonstrating}, LLMs can generate interpretable features from domain knowledge \citep{li2026human, han2024large}.
Such LLM-based AutoFE methods show promise for Data Science (DS) Agent \citep{guo2024ds,chen2025large}.
However, the assumption of access to semantic information limits their applicability in real-world scenarios \citep{nam2024optimized}.
For example, in privacy-preserving medical datasets, or sensor logs, the semantic information may be unavailable or unreliable.
In addition, the continuous expansion of the optimization trajectory used for ICL not only increases the risk of exceeding context length constraints, but also induces a bias toward successful feature-operation pairs. However, without the trajectory information, LLMs tend to generate duplicates.

To address the above challenges, we propose a {S}HAP-enhanced {I}mplicit-trajectory {G}eneration for {M}etadata-free {A}utoFE (SIGMA), a scalable constant-context optimization framework for metadata-free LLM-based AutoFE. 
Figure \ref{fig:overview} illustrates the overview of SIGMA.
First, SIGMA leverages SHAP (SHapley Additive exPlanations) values \citep{ponce2024practical} to provide task-aware signals for group feature generation.
It enables effective optimization without relying on semantic descriptions.
Specifically, input features will be divided into three groups (top, useful, and weak) according to SHAP values and used for generating intra-group and cross-group features.
Motivated by the fact that LLMs exhibit a strong contextual bias toward provided information, and that minor prompt perturbations can significantly enhance generation diversity, we propose {EX}posed-feature {I}mplicit {T}rajectory (EXIT) to mitigate the overhead associated with expanding the optimization trajectory.
In this manner, the visible feature set serves as a proxy for the optimization history, where the trajectory is implicitly reflected in the feature composition rather than being enumerated via explicit tokens.
The contributions of our work are as follows:\\
\noindent\hspace*{2em}\textbf{1.} We propose SIGMA, a metadata-free LLM-based AutoFE framework that replaces semantic descriptions with SHAP-based importance signals and introduces a grouped generation strategy for structured feature exploration.\\
\noindent\hspace*{2em}\textbf{2.} We introduce EXIT to enable effective long-horizon optimization without explicit trajectory in the prompt, which reduces the duplicate generation rate from 37.2\% to 6.8\%.\\
\noindent\hspace*{2em}\textbf{3.} Results show that SIGMA achieves comparable performance to current LLM-based baselines, and remains competitive with traditional AutoFE with efficient feature utilization.

\section{Related Work}
\textbf{Traditional AutoFE Methods.}
Traditional AutoFE methods are usually based on an expansion-reduction framework.
Deep Feature Synthesis (DFS) \citep{kanter2015deep} leveraged relational paths and mathematical primitives to automatically generate cross-table features, and then select the most performing ones.
ExploreKit \citep{katz2016explorekit} proposed a framework to generate candidate features by combining all original features, do selection on a ranking classifier.
Since non-linear transformation is also very beneficial, AutoFeat \citep{horn2019autofeat} introduced non-linear feature transformations and employs an L1-regularized linear model for feature selection, effectively enhancing the predictive power of linear models while preserving interpretability.
Besides, evolutionary computation \citep{back1996evolutionary} and genetic programming \citep{espejo2009survey} have also been widely applied in traditional AutoFE.
TPOT \citep{olson2016tpot} utilized genetic programming to combine feature selectors, transformers, and classifiers to maximize predictive accuracy.
Other representative methods include AutoGluon \citep{erickson2020autogluon}, which treats feature interactions implicitly within its hierarchical intelligent type inference and multi-stage model stacking architecture, and OpenFE \citep{zhang2023openfe}, which proposes a two-stage pruning strategy to efficiently identify high-quality candidate features.

\textbf{LLM-based AutoFE Methods.}
LLMs built on the Transformer architecture \citep{vaswani2017attention} have shown powerful ICL and reasoning capabilities \citep{guo2025deepseek}, aligned with domain knowledge.
CAAFE \citep{hollmann2023large} first proposed to take advantage of the prior semantic knowledge of LLMs to generate interpretable features based on textual descriptions.
FeatLLM \citep{han2024large} utilized LLMs to generate rules to transfer features to binary sequences based on feature descriptions and samples, boosting few-shot tabular learning.
In addition, such a framework can be easily combined with optimization algorithms.
For example, LLM-FE \citep{abhyankar2025llm} combines LLM-based feature engineering with evolutionary computation.
However, in the real-world, feature and task descriptions may be hard to obtain because of privacy and security issues, while the expansion of features will dramatically increase the prompt length.
Therefore, OCTree \citep{nam2024optimized} proposed using only the tree expression of features to generate new ones.
Although such research has achieved great success, there are several concerns about LLMs' memory of datasets \citep{zhang2024elf}, as well as the preference for generating simple operations \citep{kuken2024large}.
As a solution, \cite{li2026human} proposed decoupling the transformation operation proposal from the selection processes.

\section{Methodology}
SIGMA contains three main steps: (1) SHAP-based Feature Grouping, 
(2) Intra-Group and Cross-Group Generation, and
(3) Applying EXIT to control context size.
The overall procedure of SIGMA is summarized in Algorithm~\ref{alg:sigma}.

\vspace{0.4em}

\begin{algorithm}[!ht]
\caption{SIGMA Workflow}
\label{alg:sigma}

\begin{algorithmic}[1]
\REQUIRE Splits $(\mathcal{D}_\mathrm{tr},\mathcal{D}_\mathrm{va},\mathcal{D}_\mathrm{te})$; 
LLM $\mathcal{L}$; classifier $\mathcal{C}$; max steps $T$; 
noise level $\eta$; patience $P$

\STATE Extract $(\mathbf{X}_\mathrm{tr},\mathbf{X}_\mathrm{va},\mathbf{X}_\mathrm{te})$ from $(\mathcal{D}_\mathrm{tr},\mathcal{D}_\mathrm{va},\mathcal{D}_\mathrm{te})$
\STATE Set $s^* \leftarrow \mathrm{Eval}(\mathcal{C}, \mathbf{X}_\mathrm{tr}, \mathbf{X}_\mathrm{va})$, $\mathcal{B}\leftarrow\emptyset$, $\mathcal{H}_\mathrm{op}\leftarrow\emptyset$ and $k_\mathrm{fail}\leftarrow0$

\FOR{$t \leftarrow 1$ to $T$}
  \STATE $(\mathcal{G}_\mathrm{top}, \mathcal{G}_\mathrm{use}, \mathcal{G}_\mathrm{weak})
  \leftarrow $ SHAP-based feature grouping with $(\mathcal{C}, \mathbf{X}_\mathrm{tr},\eta)$

  \IF{$k_\mathrm{fail} \geq P$}
    \STATE Set $\mathcal{B} \leftarrow \emptyset$, $k_\mathrm{fail} \leftarrow 0$
  \ELSE
    \STATE Remove expired masks from $\mathcal{B}$
  \ENDIF

  \STATE Remove masked features in $\mathcal{B}$ from
  $(\mathcal{G}_\mathrm{top}, \mathcal{G}_\mathrm{use}, \mathcal{G}_\mathrm{weak})$

  \STATE $\mathcal{P}_t \leftarrow
  \mathrm{BuildPrompt}(\mathcal{G}, \mathcal{H}_\mathrm{op}, s^*)$


  \STATE $\Phi_t \leftarrow$ Intra/cross-group generations with $\bigl(\mathcal{L}(\mathcal{P}_t),
  (\mathbf{X}_\mathrm{tr}, \mathbf{X}_\mathrm{va}, \mathbf{X}_\mathrm{te})\bigr)$

  \STATE $\mathcal{A} \leftarrow \emptyset$

  \FORALL{$\phi \in \Phi_t$}
    \STATE $s_\phi \leftarrow \mathrm{Eval}
    \bigl(\mathcal{C},
    \operatorname{Concat}[\mathbf{X}_\mathrm{tr}, \phi(\mathbf{X}_\mathrm{tr})],
    \operatorname{Concat}[\mathbf{X}_\mathrm{va}, \phi(\mathbf{X}_\mathrm{va})]\bigr)$

    \IF{$s_\phi > s^*$}
      \STATE $\mathcal{A} \leftarrow \mathcal{A} \cup \{(\phi, s_\phi)\}$
    \ENDIF
  \ENDFOR

  \IF{$\mathcal{A} \neq \emptyset$}
    \STATE $(\phi^*, s^*) \leftarrow
    \arg\max_{(\phi, s_\phi) \in \mathcal{A}} s_\phi$ and $\mathbf{X} \leftarrow
    \operatorname{Concat}[\mathbf{X}, \phi^*(\mathbf{X})]$ for $\mathbf{X} \in \{ \mathbf{X}_\mathrm{tr},\mathbf{X}_\mathrm{va},\mathbf{X}_\mathrm{te} \}$
    \STATE Set $\mathcal{B} \leftarrow \mathcal{B} \cup
    \mathrm{Mask}(\mathrm{ExtractSourceFeatures}(\phi^*))$ and $k_\mathrm{fail} \leftarrow 0$
  \ELSE
    \STATE Set $\mathcal{H}_\mathrm{op} \leftarrow
    \mathcal{H}_\mathrm{op} \cup \mathrm{ExtractOperations}(\Phi_t)$ and $k_\mathrm{fail} \leftarrow k_\mathrm{fail} + 1$
  \ENDIF
\ENDFOR

\RETURN $(\mathbf{X}_\mathrm{tr}, \mathbf{X}_\mathrm{va}, \mathbf{X}_\mathrm{te})$ \tcp*{return augmented feature sets}

\end{algorithmic}
\end{algorithm}

\subsection{Feature Groups}
Original features are divided into groups according to their SHAP values.
Instead of dividing each group by a fixed ratio (e.g., 33\% features for each group), we introduce a threshold by adding a noise feature.
Specifically, we first add a Gaussian noise column to the original feature set, and then calculate the SHAP values for all of them.
To make features comparable to the noise, min-max normalization is applied at this stage.
After that, the SHAP value of the noise is used as the threshold to divide features into three groups: top, useful, and weak.
The top group contains features with SHAP values ranked as the top 10\% (a minimum of two features).
These features are the most important for the model's predictions.
The weak group consists of features with SHAP values below the threshold, while the remaining features are assigned to the useful group.

Dividing features into groups based on the noise threshold can help adjust the LLM's attention to the current feature space.
If we use a fixed partition ratio, the LLM will allocate a fixed level of attention to each group across all datasets.
However, different datasets have their own characteristics.
For instance, in some cases, the majority of features exhibit lower importance than the noise, whereas in others, all features are more important than the noise.
The more features a group contains, the more attention the LLM pays to it.
As a result, introducing the noise feature can help the LLM generate features that align with the intrinsic characteristics of the target dataset.
In addition, the group strategy can also benefit the EXIT strategy, which will be introduced later, making it indispensable.

\subsection{Intra-Group and Cross-Group Generation}
After dividing the features into groups, we build the prompt for the LLM to generate new features.
The prompt template can be found in the Appendix \ref{app:appendix_prompt}.
Instead of only generating one feature in each step, we require the LLM to generate one \textbf{intra-group} feature and one \textbf{cross-group} feature.
The intra-group feature focuses on deep feature transformation or multi-feature interaction to find the hidden patterns in the same group.
By generating intra-group features, we aim to improve already influential signals.
In contrast, the cross-group feature focuses on synergy building by bridging high-importance features with weaker signals.
In other words, the cross-group generation is designed to improve weak signals.
Both generated features will be evaluated by the downstream model, and only the feature with the most positive improvement will be accepted. 

\subsection{{EX}posed-feature {I}mplicit {T}rajectory (EXIT)}
Since we drop the explicit trajectory in the prompt, the LLM demonstrates a high tendency to generate duplicated features.
Motivated by the principle that information is conveyed not only through the presence of explicit signals but also through their strategic omission, we propose EXIT.
Specifically, at each step, all selected features $f_{\mathrm{sel}}$ used for generation are tracked and masked from the prompt in the following few steps.
Since top features carry more information and are more likely to benefit from interactions, they are assigned the shortest masking interval of $2^0$ steps.
In contrast, the intervals for useful and weak features are set to $2^1$ and $2^2$ steps, respectively, corresponding to their importance and function.
Given that the LLM may become trapped in local optima, EXIT resets the masked feature space and restores all frozen features when no features are accepted for $P=5$ consecutive steps.

We also track the applied operations and prevent the LLM from using the two most frequent operations in the prompt, since the LLM tends to select the same operations for a given dataset.
Nevertheless, only operations associated with failed generations are tracked and temporarily forbidden.
If an operation continuously improves performance, it should be considered well-suited to the dataset and rewarded accordingly.

In a word, the trajectory information is implicitly encapsulated within the set of exposed features, rather than maintaining an explicit, token-heavy record of the optimization history.

\section{Experiments and Results}
\subsection{Experimental Setup}
\noindent\textbf{Datasets:} 
In our experiments, we used 16 public tabular classification datasets from previous studies \citep{hollmann2023large,nam2024optimized}. 
They all come from OpenML \citep{feurer2021openml} and Kaggle \citep{banachewicz2022kaggle}.
We limited each dataset to a maximum of 50,000 samples and split it into training and test sets with an 8:2 ratio.
In addition, we performed this data splitting three times using different random seeds to improve the reliability of the results.

\noindent\textbf{Evaluation Metrics:} 
We adopt the F1-score as our primary evaluation metric to provide a balanced assessment of model performance. 

\noindent\textbf{Baselines:} 
We compare SIGMA with both LLM-based and traditional AutoFE approaches and choose XGBoost \citep{chen2016xgboost} as the downstream model.
%
As representative LLM-based approaches, we used CAAFE and OCTree as baselines.
While CAAFE is a semantic-based approach leveraging detailed feature descriptions, OCTree uses tree-structured expressions of feature space as trajectory information to realize non-semantic generation. 
%
We used well-known AutoFE methods, including DFS, OpenFE, and AutoFeat, as traditional baselines. 
These approaches are based on predefined transformation rules and perform feature generation through an exhaustive search over operation spaces.

\noindent\textbf{Experimental Protocol:}
To ensure a comprehensive evaluation, we adopt the following settings.
For SIGMA and traditional feature engineering approaches, semantic information was removed by masking feature names and encoding values, so that all methods operate without access to semantic descriptions. 
To eliminate differences arising from transformation definitions, we further adopted a shared operation space consisting of basic arithmetic operations (addition, subtraction, multiplication, division), common unary transformations (logarithm, square root, absolute value), and simple feature interactions (ratios).

For LLM-based baselines, we followed their original implementations.
To reduce performance fluctuations caused by the temperature parameter and sampling strategies, we repeated each LLM-based AutoFE method three times.
The generation budget was set to 50 features, counting generated features rather than accepted features.

We argue that the performance of an AutoFE method should also be measured by the trade-off between predictive performance and feature budget.
In the real world, a large number of generated features will be hard to interpret and require huge maintenance costs.
Therefore, we also contrast SIGMA's feature-efficiency against traditional AutoFE by varying their feature budget $K$.


\noindent\textbf{Implementation Details:}
Given the practical usage, LLMs were deployed through vLLM \citep{kwon2023efficient} to enable efficient and scalable inference.
For all baselines, we used their official implementations with standard configurations.  
To eliminate the CPU bottleneck, we used the GPU version of XGBoost \citep{mitchell2017accelerating}.
Detailed configurations of baselines are provided in Appendix~\ref{app:imple_detail}.

\begin{table*}[t]
\centering
\footnotesize
\setlength{\tabcolsep}{12pt}
\caption{F1-score comparison of LLM-based AutoFE using Qwen3-4B-Instruct-2507.
The best results are highlighted in \textbf{bold}, and the second-best results are \underline{underlined}.
}
\begin{tabular}{l |cccc}
\toprule
Dataset & \makecell{Baseline \\ (w.o. AutoFE)} & CAAFE & OCTree & \textbf{SIGMA (ours)} \\
\midrule
eucalyptus & 64.94 $\pm$ {\tiny 1.70} & \underline{65.27 $\pm$ {\tiny 2.04}} & 65.12 $\pm$ {\tiny 1.24} & \textbf{66.39 $\pm$ {\tiny 2.29}} \\
diabetes & 73.55 $\pm$ {\tiny 5.40} & 73.58 $\pm$ {\tiny 3.62} & \underline{73.62 $\pm$ {\tiny 4.37}} & \textbf{74.82 $\pm$ {\tiny 3.16}} \\
credit-g & 74.28 $\pm$ {\tiny 3.11} & 73.46 $\pm$ {\tiny 2.90} & \underline{74.56 $\pm$ {\tiny 1.73}} & \textbf{74.84 $\pm$ {\tiny 2.19}} \\
pc1 & \underline{92.71 $\pm$ {\tiny 1.21}} & 92.64 $\pm$ {\tiny 0.95} & \textbf{92.83 $\pm$ {\tiny 1.42}} & 92.22 $\pm$ {\tiny 1.06} \\
cmc & 51.22 $\pm$ {\tiny 2.60} & \textbf{51.76 $\pm$ {\tiny 2.01}} & 49.97 $\pm$ {\tiny 1.95} & \underline{51.46 $\pm$ {\tiny 2.50}} \\
wine & \textbf{80.53 $\pm$ {\tiny 0.39}} & 80.07 $\pm$ {\tiny 1.66} & \underline{80.15 $\pm$ {\tiny 1.36}} & 79.13 $\pm$ {\tiny 0.89} \\
MagicTelescope & \underline{86.30 $\pm$ {\tiny 0.30}} & \underline{86.30 $\pm$ {\tiny 0.26}} & 86.05 $\pm$ {\tiny 0.27} & \textbf{86.58 $\pm$ {\tiny 0.46}} \\
house\_16H & \textbf{87.97 $\pm$ {\tiny 0.62}} & \underline{87.79 $\pm$ {\tiny 0.69}} & 87.75 $\pm$ {\tiny 0.60} & 87.71 $\pm$ {\tiny 0.36} \\
compass & 75.07 $\pm$ {\tiny 0.23} & \underline{76.66 $\pm$ {\tiny 0.50}} & 74.13 $\pm$ {\tiny 0.41} & \textbf{77.97 $\pm$ {\tiny 1.00}} \\
electricity & 90.43 $\pm$ {\tiny 0.20} & \underline{90.51 $\pm$ {\tiny 0.26}} & 90.33 $\pm$ {\tiny 0.20} & \textbf{90.77 $\pm$ {\tiny 0.31}} \\
jungle\_chess & 86.89 $\pm$ {\tiny 0.11} & \textbf{95.03 $\pm$ {\tiny 2.93}} & 88.22 $\pm$ {\tiny 1.13} & \underline{92.62 $\pm$ {\tiny 2.12}} \\
airlines & \underline{63.43 $\pm$ {\tiny 0.60}} & 63.35 $\pm$ {\tiny 0.25} & \textbf{63.72 $\pm$ {\tiny 0.80}} & 63.31 $\pm$ {\tiny 0.53} \\
jannis & 78.60 $\pm$ {\tiny 0.52} & \underline{78.77 $\pm$ {\tiny 0.55}} & 78.69 $\pm$ {\tiny 0.39} & \textbf{78.84 $\pm$ {\tiny 0.26}} \\
MiniBooNE & \textbf{94.09 $\pm$ {\tiny 0.44}} & 93.90 $\pm$ {\tiny 0.42} & 93.91 $\pm$ {\tiny 0.46} & \underline{93.92 $\pm$ {\tiny 0.49}} \\
road-safety & 77.90 $\pm$ {\tiny 0.65} & \textbf{79.52 $\pm$ {\tiny 0.70}} & \underline{77.96 $\pm$ {\tiny 0.65}} & 77.94 $\pm$ {\tiny 0.46} \\
covertype & 87.46 $\pm$ {\tiny 0.22} & \underline{87.79 $\pm$ {\tiny 0.36}} & 87.31 $\pm$ {\tiny 0.15} & \textbf{88.29 $\pm$ {\tiny 0.43}} \\
\midrule
Average & 79.09 & \underline{79.78 $\pm$ {\tiny 0.18}} & 79.02 $\pm$ {\tiny 0.31} & \textbf{79.80 $\pm$ {\tiny 0.23}} \\
Avg Rank & 2.75 & \underline{2.31} & 2.81 & \textbf{2.06} \\
\bottomrule
\end{tabular}
\label{tab:llm_comparison}
\end{table*}

\subsection{Comparison with Existing LLM-Based Methods}
Table~\ref{tab:llm_comparison} shows the comparison of all LLM-based methods.
Results demonstrate that SIGMA achieves competitive performance with semantic-based CAAFE. 
This indicates that LLM-based AutoFEs can generate effective features without relying on detailed descriptions.
Compared to the same metadata-free OCTree, SIGMA consistently achieves better performance across most datasets.
Results of other metrics can be found in Appendix \ref{app:addition_results}.

\begin{figure}[t]
\centering
\subfigure[]{
  \includegraphics[height=5cm,keepaspectratio]{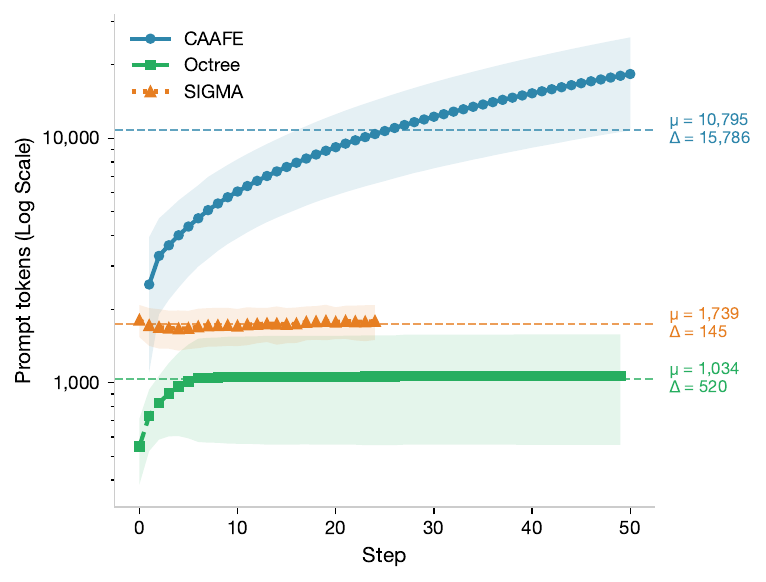}
  \label{fig:token_trend}
}
\hfill
\subfigure[]{
  \includegraphics[height=5cm,keepaspectratio]{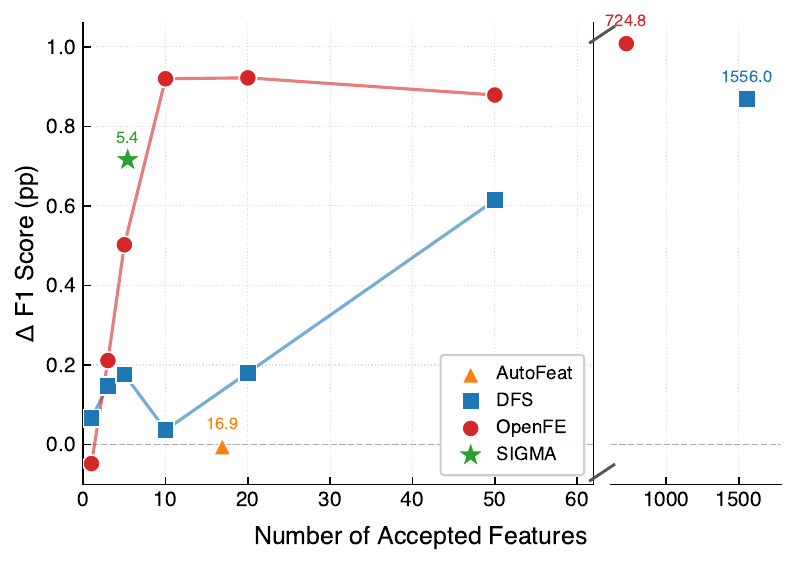}
  \label{fig:tradeoff}
}
\caption{Efficiency analysis of methods: 
    (a) the average prompt token trends of LLM-based AutoFE methods. Since SIGMA generates two features at each iteration, the final step is 25.
    (b) the feature-efficiency comparison with traditional AutoFE.}
\label{fig:efficiency_analysis}
\end{figure}

Furthermore, we analyze the prompt token trend during the optimization to compare the efficiency.
Figure~\ref{fig:token_trend} shows the average token trend of different LLM-based AutoFE methods.
We observe that both prior LLM-based AutoFE frameworks exhibit an escalating trend in prompt length across successive optimization steps.
The difference is that OCTree sets a trajectory limitation of the top-7 performing features, while CAAFE does not set any upper bound.
In contrast, SIGMA maintains a near-constant context throughout the iteration, as evidenced by the smallest peak-to-trough token variation.
While this bottleneck restricts prior methods to a limited optimization horizon, SIGMA facilitates sustainable iterative refinement without cost explosion.

By extension, although OCTree costs the fewest tokens, the optimization progress can easily become stuck when the LLM cannot generate features better than the top-7 performing features.
The absence of prompt evolution forces the system to rely solely on stochastic sampling parameters to escape the local optima.
This is one of the reasons why OCTree performs below average.
SIGMA replaces the passive strategy dependent on the LLM itself to active guidance.
Specifically, EXIT dynamically adjusts the exposed features for LLMs to reduce the co-occurrence probability of identical features.

Overall, it is observed that LLM-based AutoFE can still achieve strong performance even when semantic information is absent.
By providing the implicit trajectory through exposed features, EXIT helps SIGMA maintain the constant context during iteration, enabling long-horizon optimization.

\begin{table*}[t]
\centering\scriptsize
\setlength{\tabcolsep}{8pt}
\caption{F1-score results of traditional AutoFE methods under the same feature budget. \textbf{Bold} and \underline{underlined} indicate the best and second-best results, respectively.
}
\begin{tabular}{l |ccccc}
\toprule
Dataset & \makecell{Baseline \\(w.o. AutoFE)} & AutoFeat & DFS & OpenFE & \textbf{SIGMA (Ours)} \\
\midrule
eucalyptus & 64.94 $\pm$ {\tiny 1.70} & 65.72 $\pm$ {\tiny 2.93} & \textbf{66.71 $\pm$ {\tiny 1.71}} & 65.67 $\pm$ {\tiny 0.41} & \underline{66.39 $\pm$ {\tiny 2.29}} \\
diabetes & 73.55 $\pm$ {\tiny 5.40} & \underline{74.87 $\pm$ {\tiny 2.34}} & 74.32 $\pm$ {\tiny 2.95} & \textbf{75.93 $\pm$ {\tiny 2.82}} & 74.82 $\pm$ {\tiny 3.16} \\
credit-g & \underline{74.28 $\pm$ {\tiny 3.11}} & 73.97 $\pm$ {\tiny 2.33} & 74.18 $\pm$ {\tiny 3.05} & 73.79 $\pm$ {\tiny 2.24} & \textbf{74.84 $\pm$ {\tiny 2.19}} \\
pc1 & \underline{92.71 $\pm$ {\tiny 1.21}} & \textbf{92.94 $\pm$ {\tiny 0.66}} & 92.65 $\pm$ {\tiny 1.29} & 92.39 $\pm$ {\tiny 0.38} & 92.22 $\pm$ {\tiny 1.06} \\
cmc & 51.22 $\pm$ {\tiny 2.60} & 50.51 $\pm$ {\tiny 2.75} & \underline{52.32 $\pm$ {\tiny 2.96}} & \textbf{52.60 $\pm$ {\tiny 2.30}} & 51.46 $\pm$ {\tiny 2.50} \\
wine & \textbf{80.53 $\pm$ {\tiny 0.39}} & 78.42 $\pm$ {\tiny 1.46} & \underline{79.87 $\pm$ {\tiny 1.52}} & 78.83 $\pm$ {\tiny 0.32} & 79.13 $\pm$ {\tiny 0.89} \\
MagicTelescope & 86.30 $\pm$ {\tiny 0.30} & \underline{86.96 $\pm$ {\tiny 0.53}} & 86.05 $\pm$ {\tiny 0.10} & \textbf{87.51 $\pm$ {\tiny 0.52}} & 86.58 $\pm$ {\tiny 0.46} \\
house\_16H & 87.97 $\pm$ {\tiny 0.62} & \underline{88.04 $\pm$ {\tiny 0.53}} & 88.02 $\pm$ {\tiny 0.77} & \textbf{88.06 $\pm$ {\tiny 0.24}} & 87.71 $\pm$ {\tiny 0.36} \\
compass & 75.07 $\pm$ {\tiny 0.23} & 74.82 $\pm$ {\tiny 0.09} & 74.36 $\pm$ {\tiny 0.71} & \underline{77.27 $\pm$ {\tiny 0.75}} & \textbf{77.97 $\pm$ {\tiny 1.00}} \\
electricity & 90.43 $\pm$ {\tiny 0.20} & 90.40 $\pm$ {\tiny 0.20} & 89.97 $\pm$ {\tiny 0.13} & \textbf{91.69 $\pm$ {\tiny 0.14}} & \underline{90.77 $\pm$ {\tiny 0.31}} \\
jungle\_chess & 86.89 $\pm$ {\tiny 0.11} & 87.18 $\pm$ {\tiny 0.09} & 87.87 $\pm$ {\tiny 0.38} & \underline{90.41 $\pm$ {\tiny 0.29}} & \textbf{92.62 $\pm$ {\tiny 2.12}} \\
airlines & \textbf{63.43 $\pm$ {\tiny 0.60}} & 63.03 $\pm$ {\tiny 0.93} & 63.10 $\pm$ {\tiny 0.80} & 63.10 $\pm$ {\tiny 0.55} & \underline{63.31 $\pm$ {\tiny 0.53}} \\
jannis & 78.60 $\pm$ {\tiny 0.52} & 78.58 $\pm$ {\tiny 0.37} & \underline{78.98 $\pm$ {\tiny 0.27}} & \textbf{79.31 $\pm$ {\tiny 0.11}} & 78.84 $\pm$ {\tiny 0.26} \\
MiniBooNE & \underline{94.09 $\pm$ {\tiny 0.44}} & \textbf{94.14 $\pm$ {\tiny 0.44}} & 93.99 $\pm$ {\tiny 0.47} & 93.99 $\pm$ {\tiny 0.58} & 93.92 $\pm$ {\tiny 0.49} \\
road-safety & 77.90 $\pm$ {\tiny 0.65} & 77.87 $\pm$ {\tiny 0.38} & 77.64 $\pm$ {\tiny 0.48} & \textbf{79.70 $\pm$ {\tiny 0.68}} & \underline{77.94 $\pm$ {\tiny 0.46}} \\
covertype & 87.46 $\pm$ {\tiny 0.22} & 87.83 $\pm$ {\tiny 0.30} & 88.24 $\pm$ {\tiny 0.07} & \textbf{89.87 $\pm$ {\tiny 0.53}} & \underline{88.29 $\pm$ {\tiny 0.43}} \\
\midrule
Average & 79.09 & 79.08 & 79.27 & \textbf{80.01} & \underline{79.80 $\pm$ {\tiny 0.23}} \\
Avg Rank & 3.31 & 3.44 & 3.38 & \textbf{2.19} & \underline{2.69} \\
\bottomrule
\end{tabular}
\label{tab:traditional_comparison}
\end{table*}

\subsection{Comparison with Traditional AutoFE}
Figure~\ref{fig:tradeoff} illustrates the tradeoff between the number of selected features and the corresponding performance gain across datasets.
While DFS and OpenFE allow explicit control over the accepted feature count $K$, SIGMA and AutoFeat do not directly support feature-budget control. Therefore, each of them is shown as a single point, using the average number of generated features across datasets.
In addition, DFS and OpenFE generate an average of 724.8 and 1556 features per dataset, respectively, when all generated features are retained.
Despite their significant performance gains, these methods often introduce noisy features, rendering the underlying reasons for their effectiveness virtually uninterpretable.
Under constraint settings, OpenFE still exhibits strong feature generation abilities since it is the current most powerful AutoFE, while the performance of DFS suffers a great fluctuation.
Compared to these methods, SIGMA achieves nearly 0.8\% improvement with an average of 5.4 accepted features.
It demonstrates that SIGMA has the ability to find the most promising features under constrained settings and provides an interpretation.

In addition, Table \ref{tab:traditional_comparison} compares SIGMA and traditional AutoFE under the feature budget $K=20$, and the results of other metrics are demonstrated in Appendix \ref{app:addition_results}.
SIGMA achieves a competitive F1-score, with only a marginal gap of 0.2\% compared to OpenFE. 
Notably, this is achieved using approximately 5 accepted features, substantially fewer than the full feature budget, indicating a more efficient use of feature capacity. 
This efficiency advantage makes SIGMA a practical alternative in constrained settings.

\begin{figure}[t]
    \centering
    \subfigure[]{
        \includegraphics[height=7.5cm, trim=0cm 0cm 2cm 0cm, clip, keepaspectratio]{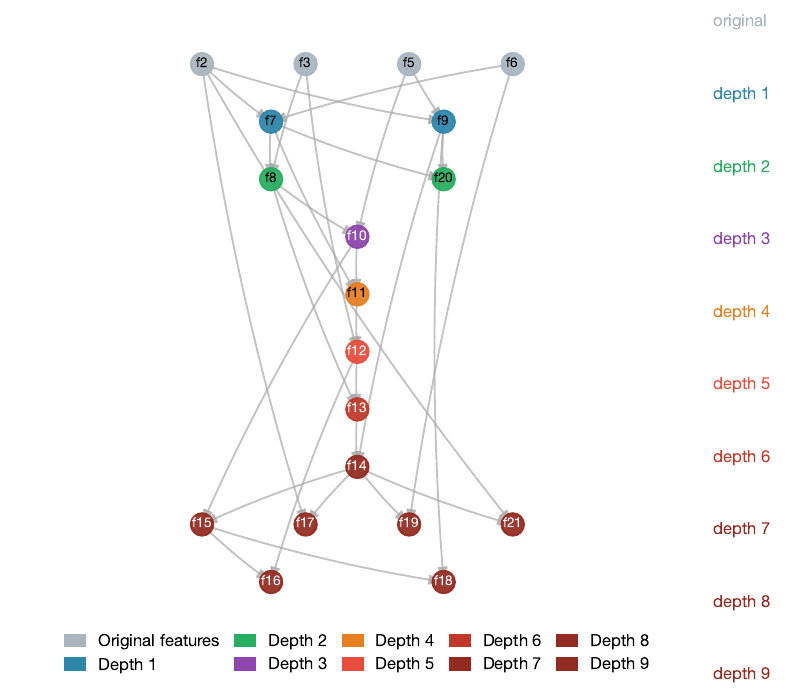}
        \label{fig:topo_jungle}
    }
    \hfill
    \subfigure[]{
        \includegraphics[height=7.5cm,trim=0cm 0cm 2cm 0cm, clip, keepaspectratio]{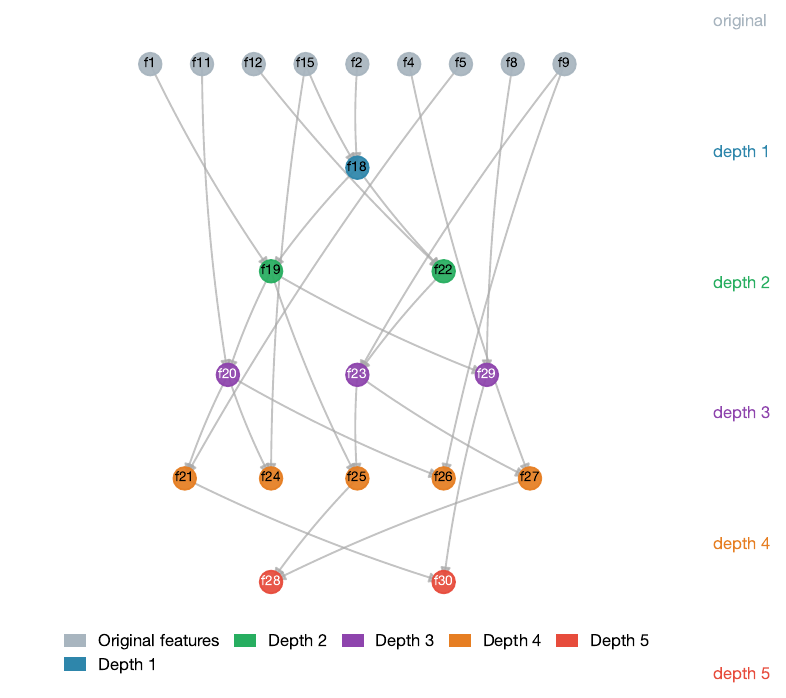}
        \label{fig:topo_campass}
    }
    \caption{The topology of generated features on the jungle\_chess and compass datasets: 
    (a) and (b) are the jungle\_chess and compass datasets, respectively.}
    \label{fig:advantage_analysis}
\end{figure}

\subsection{Case Study}
Upon reviewing the experimental results, the performance in the jungle\_chess and compass datasets are especially noteworthy due to the significant improvement.
To investigate the underlying reasons for the substantial performance gains, we performed a deeper analysis of the generated features.
Figure~\ref{fig:advantage_analysis} shows the topology structure of the generated features.
Instead of generating features independently from the original feature space, SIGMA recursively reuses previously constructed features and composes them step by step to a depth of 9, as demonstrated in Figure \ref{fig:topo_jungle}.
This is beyond the reach of traditional AutoFE methods.
A similar pattern is observed in the compass dataset, as illustrated \ref{fig:topo_campass}. 
These results highlight that SIGMA enables structured and reusable feature construction, rather than relying on shallow or independent feature generation.

\begin{figure}[t]
    \centering

    \includegraphics[width=0.5\linewidth]{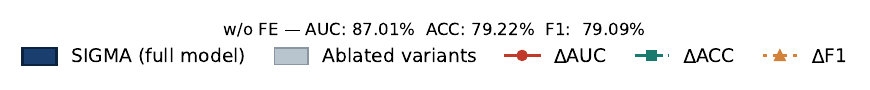}\\
    \vspace{2mm}
    \subfigure[]{
        \includegraphics[width=0.31\linewidth]{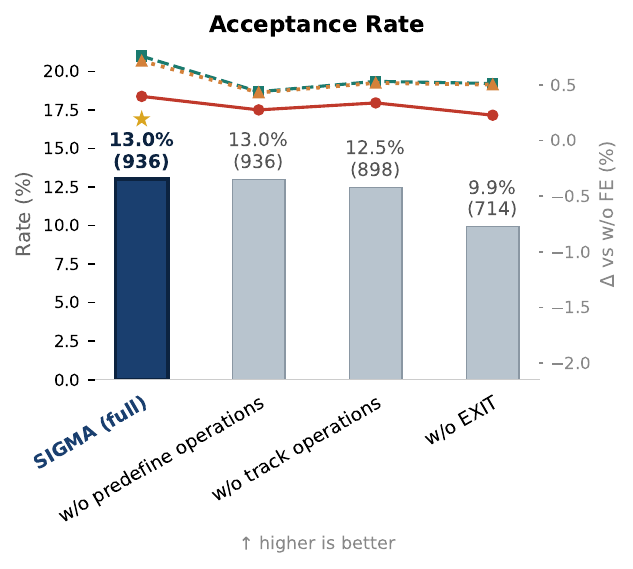}
        \label{fig:acceptance}
    }
    \hfill
    \subfigure[]{
        \includegraphics[width=0.31\linewidth]{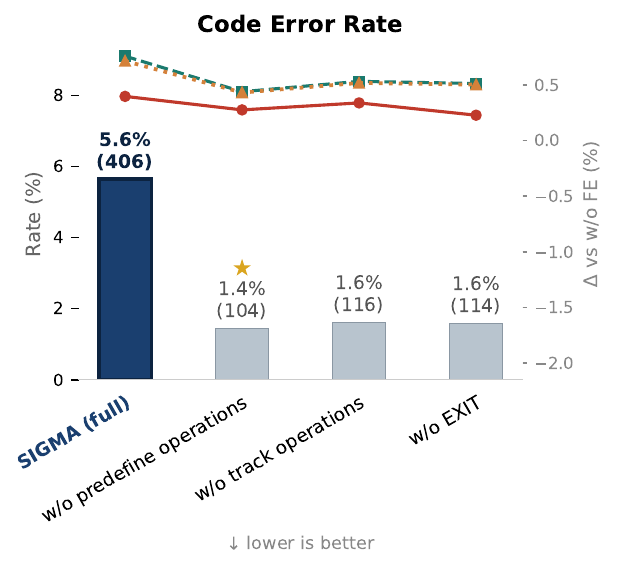}
        \label{fig:error}
    }
    \hfill
    \subfigure[]{
        \includegraphics[width=0.31\linewidth]{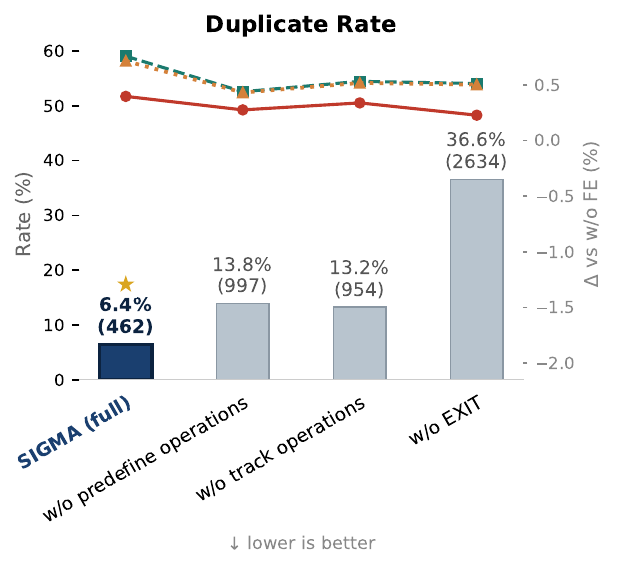}
        \label{fig:duplicate}
    }

    \caption{
    Ablation results of SIGMA.
    ``\textit{w/o predefine operations}'' represents that EXIT is still used, but operations are not provided.
    ``\textit{w/o track operations}'' represents that EXIT is still used and operations are provided, but LLMs are not forbidden from using the top-2 frequent operations.
    ``\textit{w/o EXIT}'' represents that operation predefinition and tracking are kept the same as SIGMA, but without the EXIT.
    (a) depicts the acceptance rate of generated features over all experiments.
    (b) and (c) show the code error rate and feature duplication rate, respectively.
    }
    \label{fig:ablation}
\end{figure}

\subsection{Ablation Study}
The ablation study is conducted to evaluate the effectiveness of the proposed EXIT strategies, as well as the influence on predefining and tracking operations.
Figure \ref{fig:ablation} shows the comparison between SIGMA, SIGMA without predefining operations, SIGMA without tracking the top-2 most frequent operations, and SIGMA without EXIT strategies.

Figure~\ref{fig:acceptance} shows the improvement in the acceptance rate between SIGMA and SIGMA without EXIT. 
Combined with Figure~\ref{fig:duplicate}, it can be found that without EXIT, 36.6\% of generated features are duplicated.
In other words, nearly 40\% of the generation chance has been wasted, leading to a low acceptance rate in Figure~\ref{fig:acceptance}.
EXIT successfully solves this problem as designed with a 30\% decrease in the duplicate rate.

Next, two ablated variants on operations also demonstrate another characteristic of the LLM---a tendency on operation selection.
Without predefined operations, the LLM still generates 14.1\% redundant features, a result consistent with the scenario where the top-2 most frequent operations are neither tracked nor restricted. 
This suggests an inherent heuristic bias: the LLM tends to prioritize specific operation-feature pairings based on its initial assessment, rather than exploring a broader range of alternatives.
Consequently, the optimization process remains confined to these preferred operations, leading to a low diversity in the feature generation.
This phenomenon is more serious when it is hard to generate accepted features.
The reason is that the update on feature space (accepted new features) will force the LLM to think of novel operations.
As a result, tracking and forbidding operations are also beneficial for better performance. 

However, forcing the LLM to use infrequent operations also causes a high code error rate, as shown in Figure \ref{fig:error}.
The error rate of SIGMA reaches 5.6\%, which is 4\% higher than other variants.
Therefore, when repeatedly exposed to similar feature pairs, the LLM tends to select the same operations, resulting in safer code generation but a higher duplicate rate.
This introduces a trade-off between error rate and duplicate rate.

We further analyze the impact of LLMs on SIGMA. 
We choose three representative LLMs: Qwen3-4B-Instruct (small dense), Qwen3-Coder-Next (total 80 billion parameters, activation 3 billion parameters, MoE structure), and Llama3.1-70B (large dense).
Table~\ref{tab:SIGMA_dataset_size} demonstrates the impact of different LLMs, and detailed performance results can be found in Appendix \ref{app:llms_impact}.
Although stronger models tend to achieve better average ranks across datasets, mean performance does not always improve accordingly. 
We find that this phenomenon is mainly driven by small datasets (less than 2,000 samples), where performance exhibits higher variance. 
As a result, during sequential optimization, powerful LLMs are more capable of generating features to improve the validation set performance, causing overfitting in small datasets.
In contrast, in datasets with more than 2,000 samples, larger models can continuously improve performance.
In addition, since the target task consists of coding, Qwen3-Coder-Next achieves the highest average ranking.
This overfitting on small datasets is also discovered in the field of Hyperparameter Optimization (HPO) \citep{schneider2025overtuning}, indicating that further mitigation strategies should be studied.

\begin{table}[t]
\centering
\scriptsize
\caption{SIGMA performance with different LLMs by dataset size.
LLama-70B represents Llama-3.1-70B.
The best results are highlighted in \textbf{bold}.
}
\label{tab:SIGMA_dataset_size}
\begin{tabular}{c|ccc|ccc|cc}
\toprule
\multirow{2}{*}[-2ex]{\textbf{\makecell{Dataset \\ Size}}} & \multicolumn{3}{c}{\textbf{Mean F1-Score}} & \multicolumn{3}{c}{\textbf{Mean Rank}} & \multicolumn{2}{c}{\textbf{vs Qwen3-4B}} \\
\cmidrule(lr){2-4} \cmidrule(lr){5-7} \cmidrule(lr){8-9}
& \textbf{\makecell{Qwen3\\4B}} & \textbf{\makecell{Qwen3\\Coder}} & \textbf{\makecell{Llama\\70B}} &\textbf{\makecell{Qwen3\\4B}} & \textbf{\makecell{Qwen3\\Coder}} & \textbf{\makecell{Llama\\70B}} & \textbf{\makecell{Qwen3\\Coder}} & \textbf{\makecell{Llama\\70B}} \\
\midrule
\makecell{\textbf{Small} \\ ($\leq$2000)} & \textbf{71.95\%} & 71.34\% & 71.27\% & \textbf{1.60} & 2.2 & 2.2 & -0.61\% & -0.67\% \\
\makecell{\textbf{Large} \\ ($>$2000)} & 83.37\% & \textbf{83.51\%} & 83.30\% & 2.36 & \textbf{1.73} & 1.91 & +0.14\% & -0.07\% \\
\textbf{All} & \textbf{79.80\%} & 79.71\% & 79.54\% & 2.12 & \textbf{1.88} & 2.00 & -0.10\% & -0.26\% \\
\bottomrule
\end{tabular}
\end{table}

\section{Conclusion}
In this paper, we propose SIGMA, a novel and scalable constant-context optimization framework for metadata-free LLM-based AutoFE.
Instead of semantic information, SIGMA leverages SHAP values to guide task-aware generation, and introduces a grouped generation strategy for structured feature exploration.
To enable long-horizon optimization with a low duplicate generation rate, we introduce EXIT to use exposed features in the prompt, tracking the trajectory in an implicit way.
Empirical results demonstrate that SIGMA achieves comparable performance to current LLM-based baselines with nearly constant context, and remains competitive with traditional AutoFE with efficient feature utilization.
However, several limitations remain to be addressed in future work.
First, current operation restrictions are still weak, so the duplicate rate is still nearly 7\%.
Second, SIGMA focuses only on the classification task; the regression task should also be considered.
The last is overfitting to the validation set, reflected by performance degradation.
Thus, in the future, we will focus on combining more diverse datasets and further improving performance by solving the overfitting problem with a more powerful operation selection approach.

\paragraph{Code Availability}
The source code is available at:
\url{https://github.com/shiralab/SIGMA/}

\bibliography{mybib}


\appendix

\section{Prompt Examples}
\label{app:appendix_prompt}
In Listing \ref{lst:part_1}, we give the prompt template.

\begin{lstlisting}[style=paperprompt, caption={Overall Prompt Template}, label={lst:part_1}]
You are a data science expert tasked with optimizing feature distribution to improve <CLS_MODEL> performance on a <N>-class classification problem.

## Current Data Analysis

### Feature Organization
<GROUPINGDESCRIPTION>

**Format**:
<FEATURE_FORMAT>

<FEATURE_BLOCKS>

## Your Task

### Code Generation (Required)
Generate **TWO separate Python functions** that each create **ONE new feature**:
<OPERATIONS_INFO>

#### Function 1: <FUNCTION_1_TITLE>
**Purpose**: <FUNCTION_1_PURPOSE>

**Requirements**:
<FUNCTION_1_REQUIREMENTS>

#### Function 2: <FUNCTION_2_TITLE>
**Purpose**: <FUNCTION_2_PURPOSE>

**Requirements**:
<FUNCTION_2_REQUIREMENTS>
\end{lstlisting}

\section{Implementation Details}
\label{app:imple_detail}

\textbf{AutoFeat}: The official python library is used with  \textit{feateng\_steps=2, featsel\_runs=3}

\noindent\textbf{DFS}: The official python library is used with \textit{trans\_primitives \= [`add\_numeric', `subtract\_numeric', `multiply\_numeric', `divide\_numeric', `natural\_logarithm', `square\_root', `absolute'], max\_depth=2.}

\noindent\textbf{OpenFE}: The official python library is used with default parameters.

\noindent\textbf{CAAFE}: The official Python implementation is used with XGBoost for fair comparison.

\noindent\textbf{OCTree}: The official python code is used.

\section{Additional Comparison Results of Different Metrics}
\label{app:addition_results}

Tables \ref{tab:acc_llm} and \ref{tab:acc_tradition} demonstrate the overall comparison results of LLM-based AutoFE and traditional AutoFE, respectively.

\begin{table*}[htbp]
\centering
\footnotesize
\caption{Overall accuracy (ACC) and AUC-ROC comparison of LLM-based AutoFE.}
\label{tab:acc_llm}
\begin{tabular}{l |cccc}
\toprule
Metric & \makecell{Baseline \\ (w.o. AutoFE)} & CAAFE & OCTree & SIGMA \\
\midrule
Average ACC & 79.22 & \underline{79.91 $\pm$ {\tiny 0.18}} & 79.10 $\pm$ {\tiny 0.14} & \textbf{79.98 $\pm$ {\tiny 0.24}} \\
Avg ACC Rank & 2.50 & \underline{2.38} & 3.12 & \textbf{2.00} \\
\midrule
Average AUC & 87.01 & \textbf{87.43 $\pm$ {\tiny 0.12}} & 86.77 $\pm$ {\tiny 0.10} & \underline{87.41 $\pm$ {\tiny 0.04}} \\
Avg AUC Rank & 2.44 & \underline{2.12} & 3.38 & \textbf{2.00} \\
\bottomrule
\end{tabular}
\end{table*}

\begin{table*}[htbp]
\centering\scriptsize
\footnotesize
\caption{Overall accuracy (ACC) and AUC-ROC comparison of traditional methods under a feature budget of 20.}
\label{tab:acc_tradition}
\begin{tabular}{l |ccccc}
\toprule
Dataset & \makecell{Baseline \\(w.o. AutoFE)} & AutoFeat & DFS & OpenFE & SIGMA \\
\midrule
Average ACC & 79.22 & 79.36 & 79.39 & \textbf{80.18} & \underline{79.98 $\pm$ {\tiny 0.24}} \\
Avg ACC Rank & 3.38 & 3.06 & 3.44 & \textbf{2.38} & \underline{2.69} \\
\midrule
Average AUC & 87.01 & 87.30 & 87.09 & \textbf{87.60} & \underline{87.41 $\pm$ {\tiny 0.04}} \\
Avg AUC Rank & 3.38 & \underline{2.81} & 3.44 & \textbf{2.50} & \underline{2.81} \\
\bottomrule
\end{tabular}
\end{table*}

\section{Impact of LLMs on different datasets}
\label{app:llms_impact}

The following table demonstrates the influence of different LLM backbones.

\begin{table}[htbp]
\centering\scriptsize
\setlength{\tabcolsep}{10pt}
\caption{Impact of LLMs on each dataset of F1-score. $C$ denotes the number of classes, $F$ denotes the number of features, and $N$ denotes the number of samples.}
\label{tab:llm_diff_f1}
\begin{tabular}{l | ccc | ccc}
\toprule
Dataset & $C$ & $F$ & $N$ & Llama3.1-70B & Qwen3-4B & Qwen3-Coder-Next \\
\midrule
eucalyptus & 5 & 19 & 736 & 65.15 $\pm$ {\tiny 2.35} & \textbf{66.39 $\pm$ {\tiny 2.29}} & \underline{65.43 $\pm$ {\tiny 2.45}} \\
diabetes & 2 & 8 & 768 & 73.50 $\pm$ {\tiny 3.72} & \textbf{74.82 $\pm$ {\tiny 3.16}} & \underline{73.97 $\pm$ {\tiny 3.92}} \\
credit-g & 2 & 20 & 1,000 & 73.33 $\pm$ {\tiny 2.70} & \textbf{74.84 $\pm$ {\tiny 2.19}} & \underline{73.84 $\pm$ {\tiny 2.00}} \\
pc1 & 2 & 21 & 1,109 & \textbf{92.70 $\pm$ {\tiny 1.04}} & 92.22 $\pm$ {\tiny 1.06} & \underline{92.32 $\pm$ {\tiny 1.06}} \\
cmc & 3 & 9 & 1,473 & \textbf{51.66 $\pm$ {\tiny 2.65}} & \underline{51.46 $\pm$ {\tiny 2.50}} & 51.14 $\pm$ {\tiny 3.16} \\
wine & 2 & 11 & 2,554 & \textbf{79.76 $\pm$ {\tiny 1.36}} & 79.13 $\pm$ {\tiny 0.89} & \underline{79.31 $\pm$ {\tiny 1.24}} \\
MagicTelescope & 2 & 10 & 13,376 & \textbf{86.72 $\pm$ {\tiny 0.30}} & \underline{86.58 $\pm$ {\tiny 0.46}} & 86.38 $\pm$ {\tiny 0.52} \\
house\_16H & 2 & 16 & 13,488 & \textbf{87.94 $\pm$ {\tiny 0.50}} & 87.71 $\pm$ {\tiny 0.36} & \underline{87.77 $\pm$ {\tiny 0.54}} \\
compass & 2 & 17 & 16,644 & 77.01 $\pm$ {\tiny 1.29} & \textbf{77.97 $\pm$ {\tiny 1.00}} & \underline{77.46 $\pm$ {\tiny 0.92}} \\
electricity & 2 & 8 & 38,474 & \underline{90.87 $\pm$ {\tiny 0.25}} & 90.77 $\pm$ {\tiny 0.31} & \textbf{91.33 $\pm$ {\tiny 0.34}} \\
jungle\_chess  & 3 & 6 & 44,819 & 91.17 $\pm$ {\tiny 3.06} & \underline{92.62 $\pm$ {\tiny 2.12}} & \textbf{93.45 $\pm$ {\tiny 2.59}} \\
airlines & 2 & 7 & 50,000 & \underline{63.39 $\pm$ {\tiny 0.68}} & 63.31 $\pm$ {\tiny 0.53} & \textbf{63.44 $\pm$ {\tiny 0.63}} \\
covertype & 2 & 54 & 50,000 & \textbf{88.44 $\pm$ {\tiny 0.41}} & 88.29 $\pm$ {\tiny 0.43} & \underline{88.34 $\pm$ {\tiny 0.51}} \\
jannis & 2 & 54 & 50,000 & 78.63 $\pm$ {\tiny 0.56} & \textbf{78.84 $\pm$ {\tiny 0.26}} & \underline{78.76 $\pm$ {\tiny 0.57}} \\
MiniBooNE & 2 & 50 & 50,000 & 93.88 $\pm$ {\tiny 0.48} & \underline{93.92 $\pm$ {\tiny 0.49}} & \textbf{93.94 $\pm$ {\tiny 0.53}} \\
road-safety & 2 & 32 & 50,000 & \textbf{78.54 $\pm$ {\tiny 0.77}} & 77.94 $\pm$ {\tiny 0.46} & \underline{78.40 $\pm$ {\tiny 0.73}} \\
\midrule
Average &  &  &  & 79.54 $\pm$ {\tiny 0.08} & \textbf{79.80 $\pm$ {\tiny 0.23}} & \underline{79.71 $\pm$ {\tiny 0.14}} \\
Avg Rank &  &  &  & \underline{2.00} & 2.12 & \textbf{1.88} \\
\bottomrule
\end{tabular}
\end{table}
\enlargethispage{2\baselineskip}
\end{document}